\documentclass[11pt]{article}

\usepackage[final]{acl}
\usepackage{tabularx}
\usepackage{times}
\usepackage{latexsym}
\usepackage[T1]{fontenc}
\usepackage[utf8]{inputenc}
\usepackage{microtype}
\usepackage{graphicx}
\usepackage{amsmath,amssymb,array,booktabs}
\usepackage{xcolor}
\usepackage{tikz}
\usetikzlibrary{arrows.meta,calc}
\definecolor{pipeBoxFill}{HTML}{EEF2F8}%
\definecolor{pipeBoxLine}{HTML}{2B3A55}%
\definecolor{pipeText}{HTML}{14213D}%
\definecolor{pipeAccFill}{HTML}{FDF1DE}%
\definecolor{pipeAccLine}{HTML}{8A5A12}%
\definecolor{pipeNote}{HTML}{3B4A63}%
\definecolor{pipeDim}{HTML}{5B6B85}%
\makeatletter
\newcommand{\figeightpt}{\@setfontsize\figeightpt{8}{9.6}}%
\makeatother
\tikzset{
  pipebox/.style={
    draw=pipeBoxLine, line width=.5pt, fill=pipeBoxFill, rounded corners=1.5pt,
    text=pipeText, align=center, inner xsep=3pt, inner ysep=3pt,
    minimum height=35pt, anchor=north west, font=\figeightpt},
  pipeacc/.style={pipebox, draw=pipeAccLine, fill=pipeAccFill},
  pipearrow/.style={
    draw=pipeBoxLine, line width=.5pt, -{Stealth[length=3.2pt,width=3.2pt]}},
  pipetitle/.style={text=pipeText, font=\figeightpt\bfseries,
    anchor=north west, inner sep=0pt},
  pipesub/.style={text=pipeNote, font=\figeightpt,
    anchor=north west, inner sep=0pt},
  pipenote/.style={text=pipeNote, font=\figeightpt,
    anchor=north west, inner sep=0pt},
  pipedim/.style={text=pipeDim, font=\figeightpt,
    anchor=north west, inner sep=0pt},
  pipelab/.style={text=pipeNote, font=\figeightpt, anchor=east},
}%
\newcommand{\pipelinepanel}[5]{%
  \node[pipetitle] (ptitle) at (0pt,#1) {#2};%
  \node[pipesub] (psub) at (0pt,#1-11pt) {trained only on non-held-out data
    (4 inner partitions $\rightarrow$ out-of-fold predictions)};%
  \node[pipebox,text width=92pt] (r1a) at (0pt,#1-24pt)
    {Inner training groups\\(source text,\\interpreted text)};%
  \node[pipebox,text width=115pt] (r1b) at (108pt,#1-24pt)
    {Quality model $f_\theta$\\(frozen CometKiwi\\encoder, 2 rubric heads)};%
  \node[pipeacc,text width=116pt] (r1c) at (239pt,#1-24pt)
    {OOF predicted quality\\%
     $(\hat q_s^{\mathrm{LQ}},\hat q_s^{\mathrm{EXP}})$ $+$ structure $\psi(x_s)$\\%
     $+$ delay $\phi(d_s)$};%
  \node[pipebox,text width=44pt] (r1d) at (371pt,#1-24pt)
    {Ridge LAT\\model $g_\beta$};%
  \draw[pipearrow] (r1a.east) -- (r1b.west);%
  \draw[pipearrow] (r1b.east) -- (r1c.west);%
  \draw[pipearrow] (r1c.east) -- (r1d.west);%
  \draw[pipearrow] (r1d.south) -- ++(0pt,-13pt);%
  \node[pipelab] at ($(r1d.south)+(-3pt,-6.5pt)$) {apply fitted model};%
  \node[pipebox,text width=92pt] (r2a) at (0pt,#1-72pt) {#3};%
  \node[pipebox,text width=115pt] (r2b) at (108pt,#1-72pt) {#4};%
  \node[pipeacc,text width=116pt] (r2c) at (239pt,#1-72pt)
    {Outer predicted quality\\%
     $(\hat q_s^{\mathrm{LQ}},\hat q_s^{\mathrm{EXP}})$ $+$ structure $\psi(x_s)$\\%
     $+$ delay $\phi(d_s)$};%
  \node[pipebox,text width=44pt] (r2d) at (371pt,#1-72pt)
    {Fitted $g_\beta$\\$\rightarrow \hat y_s$};%
  \draw[pipearrow] (r2a.east) -- (r2b.west);%
  \draw[pipearrow] (r2b.east) -- (r2c.west);%
  \draw[pipearrow] (r2c.east) -- (r2d.west);%
  \node[pipenote] (pnote) at ($(r2a.south west)+(0pt,-4pt)$) {#5};%
  \node[pipedim] (pdim) at ($(r2a.south west)+(0pt,-15pt)$)
    {$x_s = $ (source transcript, interpreted-output transcript)};%
}%

\definecolor{segBar}{HTML}{566278}%
\definecolor{segBoxFill}{HTML}{FDF1DE}%
\definecolor{segBoxLine}{HTML}{8A5A12}%
\definecolor{segPanelFill}{HTML}{EEF2F8}%
\definecolor{segPanelLine}{HTML}{2B3A55}%
\definecolor{segText}{HTML}{14213D}%
\definecolor{segNote}{HTML}{3B4A63}%
\tikzset{
  segbar/.style={draw=segBar, line width=2.2pt, line cap=round},
  segbox/.style={draw=segBoxLine, line width=.5pt, fill=segBoxFill,
                 rounded corners=1pt},
  segtick/.style={draw=segBoxLine, line width=.6pt},
  segrow/.style={text=segText, font=\figeightpt\bfseries,
                 anchor=north west, inner sep=0pt},
  seglabl/.style={text=segNote, font=\figeightpt, anchor=north east,
                  inner sep=0pt},
  seglabr/.style={text=segNote, font=\figeightpt, anchor=south west,
                  inner sep=0pt},
  segdelaylab/.style={text=segNote, font=\figeightpt, anchor=north,
                      inner sep=0pt},
  segdslab/.style={text=segNote, font=\figeightpt, anchor=south,
                   inner sep=0pt},
  segpanel/.style={draw=segPanelLine, line width=.5pt, fill=segPanelFill,
                   rounded corners=1.5pt},
  segdef/.style={text=segText, font=\figeightpt, anchor=north west,
                 align=left, inner sep=0pt},
  segbrace/.style={draw=segPanelLine, line width=.5pt},
  segbracetick/.style={draw=segPanelLine, line width=.5pt},
}%

\title{
Modeling Perceived Promptness in Simultaneous Interpreting from Quality, Structure, and Timing Signals
}

\author{
Ziyu Zhang$^{1}$ \and Satoshi Nakamura$^{2}$ \\
$^{1}$ School of Data Science, The Chinese University of Hong Kong, Shenzhen, China \\
$^{2}$ School of Artificial Intelligence, The Chinese University of Hong Kong, Shenzhen, China \\
\texttt{ziyuzhang@cuhk.edu.cn} \quad
\texttt{snakamura@cuhk.edu.cn}
}
\begin{document}
\maketitle

\begin{abstract}
Simultaneous interpreting requires interpreters to convey the source accurately while keeping pace with it in real time. Timing measures such as segment-onset delay capture part of this requirement, but may not fully reflect how professional evaluators judge promptness. We investigate whether these judgments can be explained by timing alone or whether the interpreted output provides additional information. Our dataset contains 622 Chinese--English and English--Chinese segments from 16 source speeches and seven interpreters. Two senior professional interpreters rated each segment for language quality (LQ), expressiveness (EXP), and perceived latency (LAT), which we also refer to as professional promptness. On held-out source speeches, onset delay alone achieves a Pearson correlation of $r=.318$ with LAT ratings. Combining predicted LQ and EXP with delay improves the correlation to $r=.475$. Lexical and structural features provide a stronger signal, reaching $r=.632$ when combined with delay and $r=.658$ when predicted quality is also included. These results suggest that professional promptness judgments reflect properties of the interpreted output beyond when the interpreter begins speaking. The additional contribution of predicted quality over structural features, however, is modest. Although the two evaluators differ in how strongly quality relates to LAT, cross-evaluator predictions remain positively correlated with LAT ratings. The overall pattern also holds when each interpreter is held out in turn.

\end{abstract}

\section{Introduction}
Simultaneous interpreting (SI) requires interpreters to understand an incoming source while producing a translation in real time. This places simultaneous demands on comprehension, memory, and production~\citep{gile1995effort,seeber2011cognitive}. Interpreting quality therefore involves more than content accuracy. Professional assessments may consider several aspects of performance separately, whereas automatic evaluation of simultaneous translation and speech translation often measures quality and latency independently~\citep{han2021reimagined,ma2019stacl,ma2020simulst,kano2023atd}.

Professional interpreting assessment is not limited to physical timing and typically focuses on overall performance rather than individual segments. In this study, we use the segment-level analytic rubric introduced by \citet{zhang2026rubric}, which distinguishes three dimensions. Language quality (LQ) assesses the accuracy and completeness of the conveyed meaning, while expressiveness (EXP) concerns fluency, naturalness, and delivery. Perceived latency (LAT), which we also refer to as \emph{professional promptness}, captures how well an interpretation is judged to keep pace with the source. All three dimensions are rated on a 0--3 scale, with higher scores indicating better performance. Thus, a high LAT score indicates good perceived synchronization rather than a longer delay. This rubric provides a way to organize professional judgments at the segment level, but is not intended as a field-wide standard.

Timing can be measured in different ways. Ear--voice span (EVS), for example, measures the temporal distance between aligned source and target anchors. Previous studies have examined how EVS can be measured automatically and how it varies across texts and interpreters~\citep{guo2024manual,janikowski2025evs}. In this study, we use a simpler measure: segment-onset delay, defined as the onset time of an interpreted segment minus that of its paired source segment.

However, measured delay is not the same as a LAT rating. An interpretation may begin quickly but omit important information or break down during delivery. Another may start later but remain complete and easy to follow. This distinction matters because standard machine translation metrics may overlook phenomena common in simultaneous interpreting, such as segmentation, paraphrasing, summarization, and disfluency~\citep{wein2024barriers}.

Earlier work using this rubric predicted LQ and EXP from text but did not model LAT~\citep{zhang2026rubric}. Building on that work, we ask whether LAT can be explained by timing alone or whether the interpreted output provides additional information. We investigate this question at the segment level by comparing three types of signals: onset delay, simple lexical and structural features of the source and interpreted text, and automatically predicted language quality and expressiveness.

Our dataset contains 622 Chinese--English and English--Chinese segments from 16 source speeches and seven interpreters. Each segment was rated by two senior professional interpreters for LQ, EXP, and LAT. Because the evaluators often assigned different LAT scores to the same segment, we use the average of their ratings as our main prediction target and also analyze each evaluator separately.

Our results show that timing alone captures only part of the variation in LAT ratings. Adding predicted LQ and EXP to onset delay raises the correlation substantially, and simple lexical and structural features of the interpreted output are stronger still; predicted quality adds a further but modest gain on top of them. These findings suggest that professional LAT judgments reflect properties of the interpreted output beyond its onset timing. The two evaluators differ in how strongly quality relates to LAT, but cross-evaluator correlations remain positive, and the overall pattern also holds when each interpreter is held out in turn. Section~5 reports the correlations.

Our main contributions are:
\begin{itemize}
    \item We formulate LAT prediction as a segment-level task and distinguish professional LAT ratings from measured onset delay.
    \item We show that onset delay alone has limited predictive power, that simple lexical and structural features add substantially more, and that predicted LQ and EXP offer a smaller further gain.
    \item We examine differences between professional evaluators and find that the relationship between quality and LAT is not confined to ratings from the same evaluator.
    \item We evaluate on held-out source speeches and held-out interpreters, using cross-fitting to prevent label leakage.
\end{itemize}

\section{Related Work}
\paragraph{Timing in simultaneous interpreting and translation.}
Timing in simultaneous interpreting can be measured in several ways. Ear--voice span (EVS) measures the temporal distance between aligned source and target units, and recent work has studied both automatic EVS measurement and variation across texts and interpreters~\citep{guo2024manual,janikowski2025evs}. EVS estimates therefore depend partly on how source and target units are aligned.

In simultaneous machine translation, latency is often studied through generation policies and delay metrics. Wait-$k$ and prefix-to-prefix approaches make latency controllable during generation~\citep{ma2019stacl}, while later work extends this setting to simultaneous speech translation~\citep{ma2020simulst}. Average Token Delay (ATD) further considers output duration rather than only the timing of partial outputs~\citep{kano2023atd}.

Our work does not estimate EVS. We use the recorded onset difference between a source segment and its paired interpretation as a simple timing signal. This measure is used as an input and baseline, rather than as a validated EVS measure.

\paragraph{Cognitive accounts of simultaneous interpreting.}
Simultaneous interpreting places demands on listening, memory, and production at the same time. Gile's Effort Model describes these competing demands, while later work examines how they interact under cognitive load~\citep{gile1995effort,seeber2011cognitive,seeber2012cognitive}.

These accounts suggest that timing, fluency, and content quality may be related in professional judgments. We test these relationships empirically without assuming a particular causal mechanism.

\paragraph{Automatic quality estimation.}
Learned metrics such as COMET and CometKiwi estimate translation quality from source and target representations~\citep{rei2020comet,rei-etal-2022-cometkiwi}. However, standard MT metrics may miss phenomena that are common in simultaneous interpreting, including segmentation, paraphrasing, summarization, and disfluency~\citep{wein2024barriers}.

We use a shared encoder to predict two professional quality dimensions, language quality and expressiveness, from the source and interpreted text. These predictions are then used as inputs to LAT modeling. Our goal is not to treat LQ and EXP as fully independent factors, but to test whether predicted quality provides information beyond timing and simple structural features.

\paragraph{Variation across evaluators.}
Human ratings can reflect differences in evaluator severity, expertise, and scale use. Such variation has also been observed in interpreter assessment, including differences in rater severity and leniency~\citep{han2015rater}. More general work on crowd labeling models annotator reliability or annotator-specific transformations~\citep{whitehill2009whose,raykar2010learning,rodrigues2018deep}.

We use the average of two professional evaluators as the main target, but we also analyze the evaluators separately. This lets us test whether the relation between quality and LAT is shared across evaluators rather than driven only by one person's scoring pattern.

\paragraph{Professional interpreting assessment.}
Professional assessment can treat interpreting performance analytically rather than reducing it to a single overall score~\citep{han2021reimagined}. Such assessment is typically concerned with interpreter performance at the task or test level. Our setting is different: we use the segment-level analytic rubric of \citet{zhang2026rubric}, which separates LQ, EXP, and LAT. This lets us study whether professional LAT judgments can be predicted from observable timing and output signals.

\section{Task, Data, and Professional Rubric}
For each segment $s$, let $y_s$ denote the LAT rating, which we also call the \emph{professional promptness score}. Higher values mean that the interpretation is judged to follow the source more promptly. Let $d_s$ denote segment-onset delay in seconds. LAT is a rating on a 0--3 scale; $d_s$ is a measured quantity in seconds. We also use human ratings of language quality and expressiveness, $q_s^{\mathrm{LQ}}$ and $q_s^{\mathrm{EXP}}$, together with their automatic estimates $\hat q_s^{\mathrm{LQ}}$ and $\hat q_s^{\mathrm{EXP}}$.

We evaluate all systems using Pearson correlation, Spearman correlation, and mean squared error (MSE). Our main evaluation holds out source speeches. We also report results when each interpreter is held out in turn.

A positive correlation means that higher model predictions are associated with higher LAT ratings. It does not mean that longer delay receives a higher score.

\subsection{Data and Timing}
Of 632 candidate segments rated by both evaluators with a numeric delay, we retain the 622 whose delay falls within the analysis range of 0 to 20 seconds; the 10 excluded segments lie outside that range. These 622 segments come from 16 source speeches and seven interpreters, of which 505 are Chinese-to-English and 117 are English-to-Chinese. Each source speech and its interpretations form one speech group. Holding out a speech group therefore prevents the same source content from appearing in both training and test data.

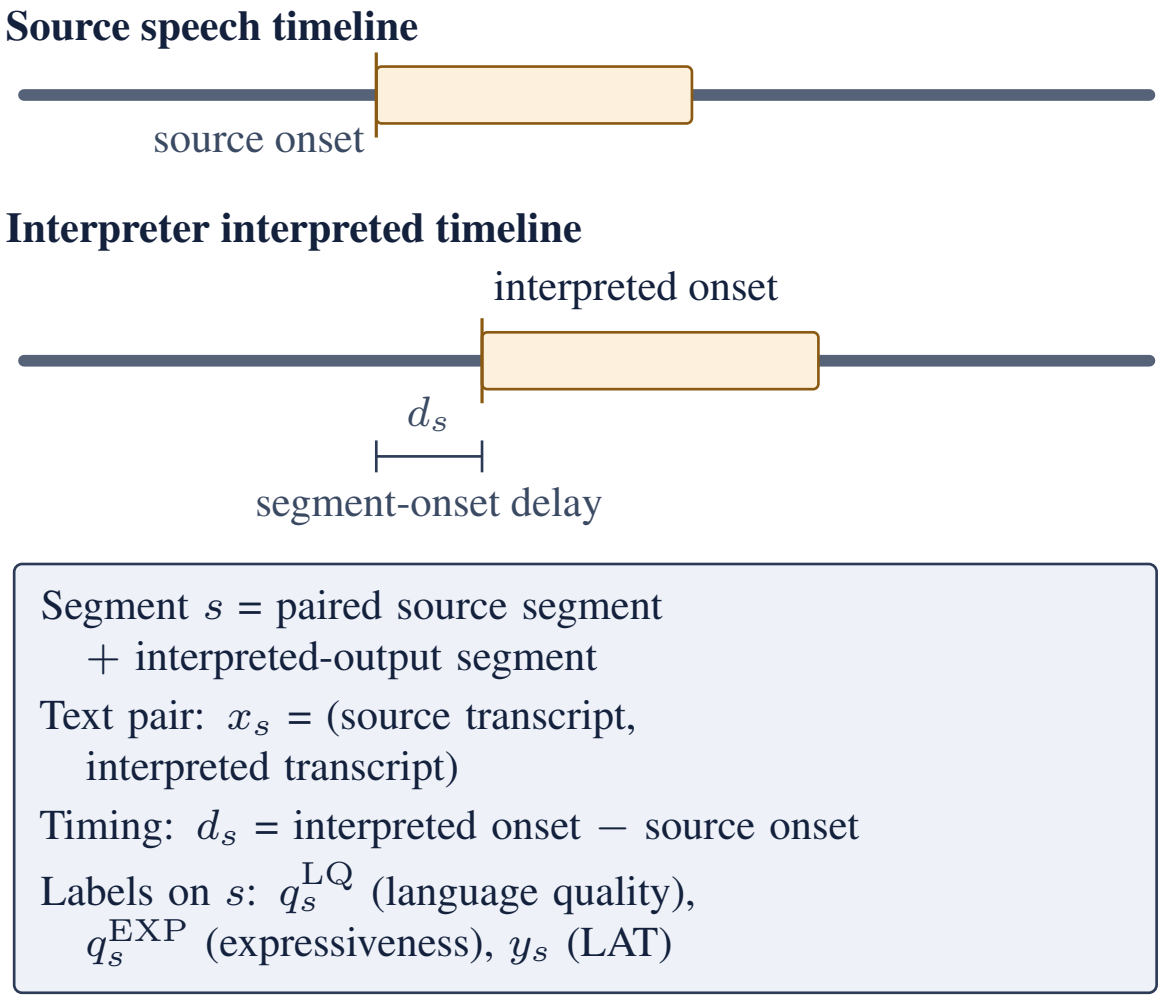
\begin{figure}[t]
\centering
%
\begin{tikzpicture}[x=1pt,y=1pt]
  \node[segrow] at (0pt,0pt) {Source speech timeline};
  \draw[segbar] (3.6pt,-16pt) -- (215.4pt,-16pt);
  \draw[segbox] (69.9pt,-10.65pt) rectangle (129.3pt,-21.35pt);
  \draw[segtick] (69.9pt,-8pt) -- (69.9pt,-24pt);
  \node[seglabl] at (67.9pt,-22pt) {source onset};
  \node[segrow] at (0pt,-38pt) {Interpreter interpreted timeline};
  \node[segdef] at (91.8pt,-49pt) {interpreted onset};
  \draw[segbar] (3.6pt,-66pt) -- (215.4pt,-66pt);
  \draw[segbox] (89.8pt,-60.65pt) rectangle (153.1pt,-71.35pt);
  \draw[segtick] (89.8pt,-58pt) -- (89.8pt,-74pt);
  \node[segdslab] at (79.85pt,-79.5pt) {$d_s$};
  \draw[segbrace] (69.9pt,-84pt) -- (89.8pt,-84pt);
  \draw[segbracetick] (69.9pt,-81pt) -- (69.9pt,-87pt);
  \draw[segbracetick] (89.8pt,-81pt) -- (89.8pt,-87pt);
  \node[segdelaylab] at (79.85pt,-89.5pt) {segment-onset delay};
  \node[segpanel,segdef,anchor=north west,inner sep=5pt,text width=205pt]
       at (1.5pt,-104pt)
    {Segment $s$ = paired source segment\\%
     \hspace*{8.5pt}$+$ interpreted-output segment\\[1.5pt]%
     Text pair: $x_s$ = (source transcript,\\%
     \hspace*{8.5pt}interpreted transcript)\\[1.5pt]%
     Timing: $d_s$ = interpreted onset $-$ source onset\\[1.5pt]%
     Labels on $s$: $q_s^{\mathrm{LQ}}$ (language quality),\\%
     \hspace*{8.5pt}$q_s^{\mathrm{EXP}}$ (expressiveness), $y_s$ (LAT)};
\end{tikzpicture}%
\caption{An analysis segment and its onset delay. Segment $s$ pairs a source-speech interval with the corresponding interpreted-output interval, and carries a text pair $x_s$ and the professional labels $q_s^{\mathrm{LQ}}$, $q_s^{\mathrm{EXP}}$, and $y_s$ (LAT).}
\label{fig:segment-def}
\end{figure}

Speech groups contain between 11 and 106 segments, with a median of 37.5. Each segment consists of a paired source and interpreted output, together with timing information and professional ratings. Figure~\ref{fig:segment-def} illustrates the unit of analysis.

Segment-onset delay is calculated as the onset time of the interpreted segment minus the onset time of its paired source segment. Across the retained segments, the mean delay is 4.59 seconds and the median is 4.0 seconds.

Of the ten excluded segments, eight have negative delays and two exceed 20 seconds. We exclude them because the archive does not record enough about the original segment boundaries to establish why these values arose, so the 0--20-second range is an analysis filter rather than a claim about acceptable interpreting behavior. Table~\ref{tab:cohort} summarizes the resulting cohort.

\begin{table}[t]
\centering
\small
\caption{Primary shared-label cohort.}
\label{tab:cohort}
\begin{tabular}{>{\raggedright\arraybackslash}p{.58\columnwidth}>{\raggedright\arraybackslash}p{.32\columnwidth}}
\toprule
Property & Value \\
\midrule
Segments / speech groups / interpreters & 622 / 16 / 7 \\
Chinese$\rightarrow$English segments & 505 \\
English$\rightarrow$Chinese segments & 117 \\
Mean LQ, EXP, LAT target range & 0.5--3.0 \\
Retained segment-onset delay & 0--20 s \\
Outer-test speech-group size & 11--106 \\
Delay-bin counts (0--2 / 2--4 / 4--6 / 6--10 / 10--20 s) & 82 / 202 / 151 / 138 / 49 \\
\bottomrule
\end{tabular}
\end{table}

Two senior professional interpreters, each with more than ten years of experience, rated the segments. Each evaluator used integer scores from 0 to 3. We keep only segments with a valid score from both evaluators and use the mean of their ratings as the main target. This produces scores in 0.5-point steps.

The two evaluators show limited agreement on LAT, though 91.5\% of their ratings differ by no more than one point. We therefore use the mean as a practical reference target and analyze the two evaluators separately in Section~5.4, which reports the agreement statistics.

\subsection{Professional Rubric and Rating Procedure}
The evaluators rated the three dimensions of the rubric of \citet{zhang2026rubric}: language quality (LQ) for the adequacy and accuracy of the conveyed content together with well-formed target-language expression, expressiveness (EXP) for fluency, naturalness, and delivery, and perceived latency (LAT) for how well the interpretation is judged to keep pace with the source. All three use a 0--3 scale on which higher values indicate better performance, so a high LAT score denotes an interpretation judged to follow the source promptly rather than one with a longer delay. Individual ratings use integer anchors and the shared mean target has .5-point increments. The complete anchors are provided for reference (see Table~\ref{tab:retained-grid} in Appendix~A).

The LAT score is not defined by a fixed timing threshold. A segment with a short onset delay can still receive a low LAT score if the interpretation loses important information or breaks down during delivery. Conversely, a longer delay does not necessarily lead to the lowest score if the interpretation remains coherent and complete.

During rating, evaluators listened to synchronized source and interpretation audio and could also view the corresponding transcripts. Replay was allowed. The numerical onset delay was not shown to them. The transcripts were provided to support judgments of content and delivery, rather than as direct evidence of timing.

All three dimensions were rated in the same session, so influence across dimensions cannot be fully ruled out. In particular, a general impression of quality may affect LAT ratings. We address this concern in Section~5.4 by testing whether quality ratings from one evaluator are also related to LAT ratings from the other.

The archive contains the source and interpreted text, onset differences, scores, and optional comments for each segment. It does not preserve detailed information about how the original segment boundaries or onset times were created. We therefore treat segment-onset delay as the recorded timing measure used in this dataset, rather than as a validated EVS estimate.

Free-text comments were optional and include observations about omissions, inaccuracies, fluency, pacing, and sentence construction. These comments are used only for qualitative analysis and are never given to the models.

\section{Modeling Professional Promptness}
\subsection{Onset Delay}
We first model LAT using segment-onset delay alone. Because the relation between delay and professional ratings may not be strictly linear, we represent delay with a piecewise-linear basis using knots at 2, 4, 6, and 10 seconds,
\begin{equation}
\begin{aligned}
\phi(d_s) = [&\, d_s,\,(d_s-2)_+,\,(d_s-4)_+,\\
             &\, (d_s-6)_+,\,(d_s-10)_+ ] ,
\end{aligned}
\end{equation}
where $(x)_+=\max(0,x)$. The knot locations follow the descriptive delay ranges used in our analysis and are not tuned on test data. We also test linear, quadratic, logarithmic, alternative-knot, and binned representations as sensitivity checks. These variants move the delay-only correlation by approximately $.06$ at most, and none of them closes the gap to models that add lexical and structural features. Changing the functional representation of delay therefore does not account for the limited predictive power of timing alone.

\subsection{Lexical and Structural Features}
We next add simple features derived from the source and interpreted text. These include source and output length, the output-to-source length ratio, punctuation and sentence-ending counts, lexical diversity, a very-short-output indicator, and language direction.

For Chinese text, each CJK character is counted as one lexical unit; for alphabetic text, we count alphanumeric words. These units are not equivalent, so length values are not directly comparable across language directions; the additive direction indicator lets the model adjust for this systematic difference between the two directions. All numeric features are standardized using only the training portion of each fold.

The very-short-output indicator fires when an output has fewer than five lexical units. No retained segment meets this condition, so the indicator is constant at zero throughout and contributes no predictive variation in our experiments. We retain this inactive feature for reproducibility (see Table~\ref{tab:family-ablation} in Appendix~E).

These features do not use human ratings. They provide a simple baseline for testing whether LAT can already be explained by observable properties such as output length, completeness, and structure.

\subsection{Predicted Quality}
We also estimate two professional quality dimensions from the source and interpreted text: language quality (LQ) and expressiveness (EXP).

The quality model uses a frozen \texttt{Unbabel/wmt22-cometkiwi-da} encoder followed by two prediction heads, one for LQ and one for EXP. The input is the ordered pair of source transcript and interpreted output. The encoder uses the implementation tokenizer with a maximum length of 512 tokens, and the two heads are trained for ten epochs with AdamW (see Appendix~B for architectural and training details).

The automatic quality estimates are denoted by $\hat q_s^{\mathrm{LQ}}$ and $\hat q_s^{\mathrm{EXP}}$. We also report results using human LQ and EXP ratings as a reference, since these ratings would not be available at inference time.

LQ and EXP predictions are strongly correlated, so we do not interpret them as fully independent latent factors. Our goal is simply to test whether predicted professional quality adds information beyond timing and simple structural features.

\subsection{Evaluation on Held-Out Source Speeches}
Our main evaluation holds out one source-speech group at a time. All interpretations of the held-out speech are excluded from training the quality model and the final LAT model.

The procedure has two stages. First, the quality model produces LQ and EXP estimates. For the training portion, these estimates are generated out of fold, so that no segment receives a quality prediction from a model trained on that segment's own quality labels. A separate quality model trained without the held-out speech then predicts LQ and EXP for the test group.

Second, a Ridge regressor predicts LAT, $\hat y_s=g_\beta([\hat q_s,\phi(d_s),\psi(x_s)])$, from one of five feature sets: onset delay only; predicted LQ and EXP only; predicted LQ and EXP plus delay; lexical and structural features plus delay; or all three signal types together, where $\psi(x_s)$ denotes the structural features. All scaling is fitted on the training data within each fold. Ridge regularization is fixed at $\alpha=1.0$. We repeat this procedure for all 16 source speeches and concatenate the held-out predictions before computing corpus-level metrics.

\begin{figure*}[t]
\centering
\begin{tikzpicture}[x=1pt,y=1pt]
  \pipelinepanel{0pt}{Held-out source speeches}
    {Held-out speech group $g$\\(source text,\\interpreted text)}
    {Outer quality model\\(excludes group $g$ from\\training/dev/checkpoint)}
    {held-out group $g$ excluded from upstream quality training
     and from second-stage Ridge fitting}
\end{tikzpicture}%
\caption{The cross-fitted two-stage protocol, shown for held-out source speeches; the held-out-interpreter protocol of Section~\ref{sec:loio} is identical with the interpreter in place of speech group~$g$. $\hat q_s^{\mathrm{LQ}}$ and $\hat q_s^{\mathrm{EXP}}$ are cross-fitted automatic quality estimates, $\psi(x_s)$ the structural features, and $\phi(d_s)$ the onset-delay basis.}
\label{fig:pipeline}
\end{figure*}
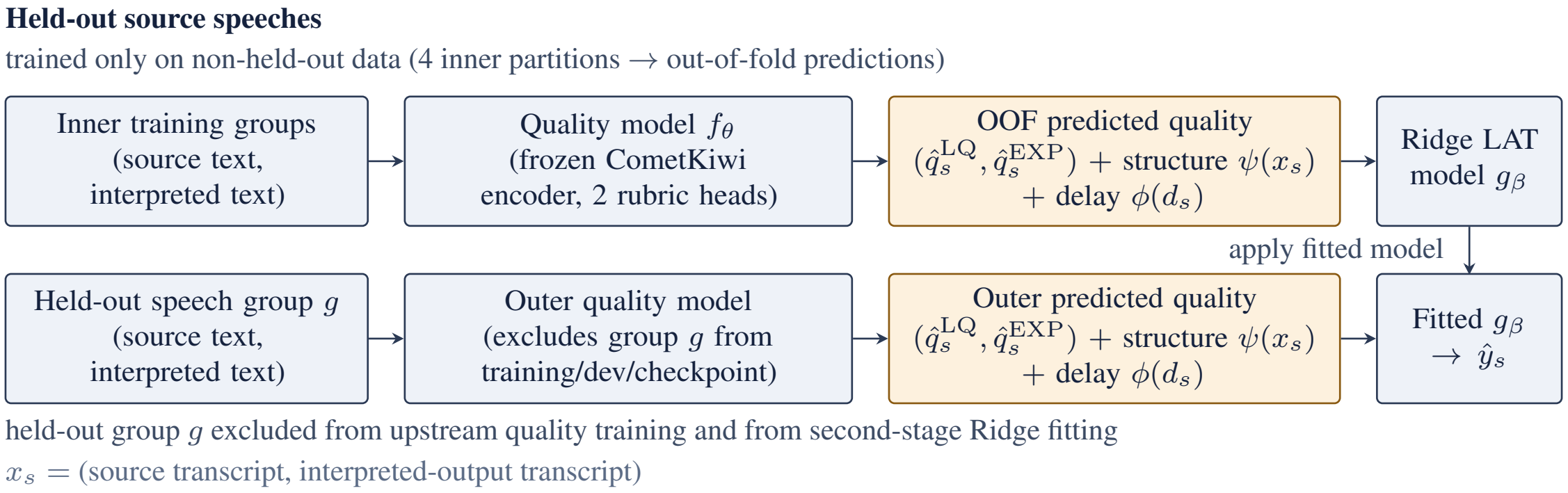

This design (Figure~\ref{fig:pipeline}) ensures that the test speech is not used for quality supervision, model selection, or LAT-model training. It also prevents the second-stage model from receiving in-sample quality predictions for its training examples. Fold construction, checkpoint selection, and other implementation details are given in Appendix~B. Because Ridge predictions are not bounded to the rating scale, we also test clipping and a bounded regression model as sensitivity analyses (Section~5.2).

\subsection{Evaluation with Held-Out Interpreters}
\label{sec:loio}
We also test whether the same pattern remains when each interpreter is held out in turn.

For a given test interpreter, all of that interpreter's segments are excluded from quality-model training, model selection, and LAT-model training. The remaining data are used to generate out-of-fold quality predictions for training, while a final model trained without the held-out interpreter produces quality estimates for the test segments.

We evaluate four of these feature sets: onset delay only, predicted LQ and EXP plus delay, lexical and structural features plus delay, and all three together. The predicted-quality-only set is omitted here, since this experiment is concerned with whether the signals that help on held-out speeches still help on held-out interpreters. Interpreter identity is not used as an input.

This setting tests whether the overall relation between timing, output properties, and LAT remains when an interpreter is held out during training. Source speeches may still appear through other interpreters, so this experiment does not test the harder case in which both the interpreter and source speech are held out.

\subsection{Statistical Analysis}
For models that depend on random initialization, we report the mean and standard deviation across three fixed seeds. The reported standard deviation reflects variation across seeds rather than sampling uncertainty.

For the main source-speech evaluation, we estimate uncertainty with 10,000 bootstrap samples. We resample the 16 source-speech groups with replacement and keep all segments within each sampled group together.

Our main statistical comparison asks whether predicted quality adds information beyond timing alone. We compare predicted LQ/EXP plus delay against delay alone using the difference in Pearson correlation,
\begin{equation}
T = r(y,\hat{y}_{Q+D}) - r(y,\hat{y}_{D}).
\end{equation}
We compute an exact two-sided permutation test by swapping the paired system predictions within whole speech groups across all $2^{16}=65{,}536$ possible assignments. We treat this as our only confirmatory test; implementation details are given in Appendix~B.

We also compare the full model with the structural-features-plus-delay model. Because this is a secondary comparison, we report its bootstrap confidence interval and seed-level results without treating it as a second confirmatory hypothesis test.

\section{Experiments and Analysis}
\subsection{Automatic Quality Prediction}
Across the source-speech evaluation folds, the automatic quality model reaches Pearson $r=.450\pm.051$ for LQ and $r=.415\pm.013$ for EXP. The predictions show substantial variation across segments rather than collapsing to a narrow range. LQ and EXP predictions are also strongly correlated ($r=.818$--$.894$), suggesting that they capture overlapping aspects of interpretation quality rather than fully separate factors.

\subsection{Main Results}
Table~\ref{tab:main} compares the main systems on held-out source speeches.

Timing alone is the weakest of the three signals. Adding predicted LQ and EXP to onset delay raises the correlation from $.318$ to $.475$, and this difference is significant under the exact speech-group permutation test, which is our only confirmatory test. Its statistic is computed from seed-averaged predictions rather than from the difference between table entries, and gives $\Delta r=.170$ ($p=.000687$).

\begin{table}[t]
\centering
\scriptsize
\setlength{\tabcolsep}{2.5pt}
\caption{Results on held-out source speeches, with Pearson $r$ and Spearman $r_S$. Pred.~Q denotes predicted LQ and EXP, and struct.\ the lexical and structural features. Learned models report mean $\pm$ standard deviation across three seeds. The human-quality reference uses human LQ and EXP ratings alone, without onset delay, and is a reference point rather than a competing system.}
\label{tab:main}
\begin{tabular}{@{}lccc@{}}
\toprule
System & $r$ & $r_S$ & MSE \\
\midrule
Delay only & .318 & .347 & .281 \\
Pred. Q & $.435\pm.037$ & $.463\pm.025$ & $.259\pm.012$ \\
Pred. Q + delay & $.475\pm.026$ & $.515\pm.016$ & $.246\pm.010$ \\
Struct. + delay & .632 & .618 & .187 \\
Pred. Q + struct. + delay & $\mathbf{.658\pm.010}$ & $\mathbf{.652\pm.007}$ & $\mathbf{.177\pm.004}$ \\
\midrule
Human-quality reference & .809 & .707 & .107 \\
\bottomrule
\end{tabular}
\setlength{\tabcolsep}{6pt}
\end{table}

Simple lexical and structural features are substantially stronger: combined with delay they reach $r=.632$, and adding predicted quality raises this to $r=.658\pm.010$.

Predicted quality therefore adds much less beyond structure and delay than it does beyond timing alone. The per-seed improvements are $.033$, $.030$, and $.014$, and the bootstrap estimate, computed from seed-averaged predictions rather than by averaging the per-seed differences, is $\Delta r=.030$ $[.002,.063]$. The results suggest a small additional predictive contribution, although its magnitude varies across seeds.

\paragraph{Speech-group uncertainty.} Bootstrapping over source speeches gives a 95\% interval of $[.381,.596]$ for the predicted-quality-plus-delay model, and its improvement over delay alone remains positive across Pearson correlation, Spearman correlation, MSE, and MAE. Adding delay to predicted LQ and EXP improves the correlation by $\Delta r=.040$, with a 95\% bootstrap interval of $[.002,.066]$; this increment is smaller than the gain from adding predicted quality to delay.

\paragraph{Range sensitivity.} Ridge predictions occasionally exceed the upper end of the rating scale. Clipping them to the scale gives $r=.482\pm.022$, and a bounded regression model gives $r=.463\pm.038$, compared with $r=.475\pm.026$ for the original predicted-quality-plus-delay model. These checks do not materially change the main conclusion (see Appendix~B for details).

\subsection{What Do Structural Features Capture?}
The strong structural baseline suggests that simple features of the interpreted output are closely related to LAT ratings: output length and sentence structure may reflect how much information an interpreter conveys and how clearly it is delivered. These features do not measure interpreting quality directly, but they may capture some aspects of it, which raises the question of whether the predicted LQ and EXP scores carry information beyond them.

To examine this, we regress predicted LQ and EXP on the structural features within each training fold and use the residuals in the full LAT model. Structural features explain only part of the predicted quality scores ($R^2=.200$--$.246$ for LQ and $R^2=.114$--$.199$ for EXP).

Replacing the original quality scores with these residuals gives almost identical results across all three seeds ($r=.665,.663,.646$), but because the model already includes structural features, this similarity is not independent evidence that predicted quality adds new information. The evidence for that comes from Table~\ref{tab:main}, where adding predicted LQ and EXP improves correlation from $.632$ to $.658$: a modest gain, but one that suggests some information beyond the structural baseline (see Appendix~C for the residualization results).

\paragraph{Which structural features carry the signal?} To locate the source of the structural advantage, we ablate the structural block by feature family within the lexical/structural-plus-delay model, which involves no seeds and no predicted quality. We report observed differences in Pearson correlation, $\Delta r$, with 95\% intervals from a bootstrap that resamples the 16 source-speech groups with replacement. These intervals are exploratory and are not the confirmatory permutation test of Section~5.2.

The largest changes involve the length-related features and the language-direction indicator. Adding source and output length to delay raises correlation from $.318$ to $.486$ ($\Delta r=+.168$ $[.052,.268]$), and removing the language-direction indicator from the full model costs $\Delta r=+.141$ $[.029,.247]$. The length ratio and lexical diversity do not change the full model detectably (see Table~\ref{tab:family-ablation} in Appendix~E).

The structural features are strongly correlated, so these are feature-group removal effects rather than estimates of individual feature importance, and no configuration here removes length information from the model entirely. The additive direction indicator plausibly helps the model adjust for systematic differences between the two language directions, but we do not interpret it as an identified direction effect: every speech group is single-direction, so direction is confounded with speech-level differences in topic, speaker, and interpreter.

\subsection{Agreement and Transfer Across Evaluators}
We also examine how the two evaluators differ in their LAT judgments. Their agreement is limited: exact agreement is 51.4\%, Pearson correlation is $.259$, and both ICC$(2,1)$ and quadratic-weighted kappa are $.236$ (see Table~\ref{tab:rater-confusion} in Appendix~C).

We then ask whether quality ratings from one evaluator can help predict LAT ratings from the other (Table~\ref{tab:cross-rater}). Using human LQ and EXP ratings together with onset delay, we obtain $r=.909$ when predicting R05's LAT from R05's quality ratings, compared with $r=.552$ for R06. When we use one evaluator's quality ratings to predict the other's LAT ratings, the correlations remain positive: $.430$ from R05 to R06 and $.543$ from R06 to R05.

\begin{table}[t]
\centering
\footnotesize
\caption{Transfer across evaluators using human LQ/EXP ratings and onset delay. Brackets show 95\% confidence intervals over source speeches.}
\label{tab:cross-rater}
\begin{tabular}{lcc}
\toprule
Quality inputs & LAT target & Pearson $r$ \\
\midrule
R05 & R05 & .909 [.855, .944] \\
R05 & R06 & .430 [.348, .499] \\
R06 & R05 & .543 [.482, .593] \\
R06 & R06 & .552 [.496, .607] \\
\bottomrule
\end{tabular}
\end{table}

\paragraph{Evaluator-specific prediction.} We also refit the delay-only and lexical/structural-plus-delay models on each evaluator's own LAT ratings, leaving the cohort, folds, features, and Ridge settings unchanged so that only the target differs. The two signals are associated with the two evaluators in different proportions. Onset delay tracks R06's ratings more closely than R05's ($r=.400$ against $.106$), whereas adding the structural features helps R05 far more, raising R05 to $.613$ ($\Delta r=+.507$ $[.329,.618]$) against R06's $.442$ ($\Delta r=+.042$ $[-.034,.123]$), an interval that includes zero (see Table~\ref{tab:rater-auto} in Appendix~E). We did not test the difference between the two improvements against each other, so the contrast is descriptive: these are predictive associations between feature sets and recorded ratings, and they do not identify what either evaluator attended to. Differences in score distribution and scale use may contribute, as R05 used all four points of the scale and R06 never used the lowest. We ran this comparison only for the onset-delay and structural feature sets; whether predicted quality behaves differently across evaluators is untested here.

Taken together with the transfer results, the association between quality and LAT holds across evaluators as well as within them, but differs considerably in strength. The mean target is a practical choice rather than a validated ground truth, and conceals these differences, so modeling evaluators separately is a reasonable next step.

\subsection{Results with Held-Out Interpreters}
In the pooled held-out-interpreter evaluation (Table~\ref{tab:loio}), the systems retain the same ordering as in the held-out-speech evaluation. Delay alone becomes much weaker ($r=.048$), and the full model again gives the highest pooled correlation ($r=.573\pm.019$).

\begin{table}[t]
\centering
\footnotesize
\caption{Results with each interpreter held out in turn. Learned models report mean $\pm$ standard deviation across three seeds; macro $r$ gives equal weight to each interpreter.}
\label{tab:loio}
\begin{tabular}{lrr}
\toprule
System & Pooled $r$ & Macro $r$ \\
\midrule
Delay & .048 & .173 \\
Pred. Q + delay & $.377\pm.080$ & $.221\pm.024$ \\
Structure + delay & .511 & .332 \\
Full model & $\mathbf{.573\pm.019}$ & $\mathbf{.358\pm.016}$ \\
\bottomrule
\end{tabular}
\end{table}

Giving each interpreter equal weight narrows the gap between the full model and structure plus delay to $.358$ against $.332$, so the relation survives holding an interpreter out, but its strength varies across individuals. Because the same source speech may still appear through other interpreters, this tests transfer to held-out interpreters, not to both at once.

\subsection{Direction-Specific Analysis}
The relative strength of the systems varies across directions and evaluators. For example, the structural-plus-delay correlation for R05 is $.508$ on Chinese-to-English segments and $.740$ on English-to-Chinese segments, while the corresponding values for R06 are $.428$ and $.466$. These differences are descriptive and cannot establish a direction effect: every speech group is single-direction, so direction is confounded with speech group, and the English-to-Chinese subset covers only six source speeches (see Table~\ref{tab:rater-direction} in Appendix~C).

\section{Conclusion}
We found that segment-onset delay alone captures only part of the variation in LAT ratings for simultaneous interpreting. Simple lexical and structural properties of the interpreted output carry substantially more predictive signal, and automatically predicted language quality and expressiveness add a smaller further gain. These associations hold across evaluators and interpreters: one evaluator's quality ratings stay positively associated with the other's LAT ratings, and system ordering is preserved in pooled held-out-interpreter results, though performance varies across interpreters. LAT is therefore better treated as a judgment reflecting both timing and output properties than as a proxy for physical delay alone; testing this will need more evaluators, more balanced language directions, and explicit models of evaluator differences.

\section*{Limitations}
This study has several limitations. First, the dataset is small and unbalanced across language directions, with only seven interpreters and two professional evaluators. The findings therefore need to be tested on broader interpreter populations, more evaluators, and more balanced language pairs.

Second, our timing measure is based on archived segment onset times rather than manually validated EVS. The archive does not preserve enough information about how segment boundaries and onset times were created to estimate timing error directly. We therefore treat onset delay as a recorded timing signal rather than as a validated perceptual latency measure.

Third, the evaluators rated LQ, EXP, and LAT in the same session while listening to synchronized audio and viewing transcripts. This setting allows careful professional judgment, but influence across dimensions cannot be ruled out. The cross-evaluator analysis reduces this concern, but does not eliminate it.

Finally, the held-out interpreter experiment does not also hold out source speeches. It therefore tests transfer to held-out interpreters, rather than the harder setting in which both the interpreter and source speech are held out. Future work should also model evaluator differences directly and test whether the same findings hold under real-time listening conditions.

\section*{Ethics Statement}
The study received approval from the relevant institutional review board, and all participating interpreters provided informed consent for the use of their recordings and associated study data.

Because the dataset contains only seven interpreters, raw recordings, verbatim transcript pairs, and the interpreter-identity crosswalk will not be released due to re-identification risk. Subject to the study's governance constraints, the planned public release covers code, anonymized professional labels, archived segment-onset-delay records, the retained rubric, deterministic fold manifests, experiment scripts, and environment specifications. Any text released with the dataset will be de-identified or redacted as required by the study's governance conditions.

The proposed models are intended for research and formative feedback. They should not be used for employment decisions, certification, or ranking interpreters. In particular, the variation we observe across evaluators and interpreters means that LAT scores should not be treated as a complete measure of interpreting quality.

\clearpage
\bibliography{references}

\clearpage
\appendix
\section*{Appendix A: Professional Rating Rubric}

Table~\ref{tab:retained-grid} gives the full rating criteria for the three dimensions summarized in Section~3.2, retained from \citet{zhang2026rubric}.

\begin{table}[h]
\centering
\scriptsize
\caption{Retained segment-level rubric definition. Individual evaluators use the integer anchors 0--3; .5-point values occur only in the two-rater mean target.}
\label{tab:retained-grid}
\begin{tabular}{@{}p{.14\columnwidth}p{.79\columnwidth}@{}}
\toprule
Dimension & Criteria summary \\
\midrule
LQ & 3: meaning is preserved with no major omission or distortion. 2: some omissions or inaccuracies, but the main message remains intact. 1: major omissions or distortions, including misleading or reversed meaning. 0: meaning is largely absent, misleading, or reversed. \\
EXP & 3: fluent, idiomatic; minor errors only. 2: generally fluent with noticeable awkwardness/errors. 1: frequent disfluency/grammar issues affecting delivery. 0: breakdown-level disfluency impedes comprehension. \\
LAT & 3: low perceived lag; well synchronized. 2: occasional lag; manageable. 1: frequent noticeable lag; disjointed flow. 0: severe persistent lag; difficult to follow. \\
\bottomrule
\end{tabular}
\end{table}

\section*{Appendix B: Implementation and Reproducibility Details}

\paragraph{Quality-model training.} The frozen \texttt{Unbabel/wmt22-cometkiwi-da} backbone encodes the (source transcript, interpreted-output transcript) pair with the implementation tokenizer (512-token truncation, first-token pooling, which takes the final-layer hidden state at the leading position of the encoded pair); two linear residual heads predict LQ and EXP. Head-only runs use AdamW for ten fixed epochs (batch size 16) with fixed weighted MSE, variance, and correlation losses; checkpoint selection maximizes development LQ Pearson+EXP Pearson. Trained head weights and the exact inference configuration will accompany the planned code release. The archive lacks transcript provenance, tokenizer revision, and token-length logs; we do not reconstruct a truncation rate. The predictor consumes complete text pairs only, so it is post-hoc rather than a streaming ASR system.

\paragraph{Fold construction for held-out source speeches.} For each held-out source-speech group $g$, all automatic predictions for $g$ come from an upstream quality model trained and selected only on other groups. Exact duplicate source-text--interpreted-output pairs from $g$ are removed from upstream train and development partitions. Within the remaining groups, four inner source-speech-group partitions generate out-of-fold $\hat q^{\mathrm{LQ}}$ and $\hat q^{\mathrm{EXP}}$ for every LAT-training row; a final outer-excluded quality model predicts group $g$. The second-stage Ridge regressor is trained on inner out-of-fold features and evaluated on the final outer prediction, so no row's second-stage quality feature is produced by a model that used that row's LQ/EXP label.

The 16 outer folds and four inner partitions are deterministic greedy segment-count balances, with stable name tie-breaks and no direction/interpreter stratification. Each final fold holds one group out; inner held-out groups supply OOF predictions, and all seeds reuse the same manifest. Quality models run for fixed ten epochs without early stopping; checkpoint selection uses development LQ Pearson+EXP Pearson, while hyperparameters are fixed and outer-test labels are never used. Ridge scaling is fit on outer LAT-training rows only. Separately, we ran a diagnostic check that replaces the LQ and EXP pair with their first principal component, fitted within each training fold, to test whether the two dimensions can be collapsed into a single quality factor. No principal-component transform is fitted or applied in any reported model; the check is a sensitivity analysis only. Outer test inputs are used solely to remove exact duplicate source--output pairs from upstream train/dev, never their labels. Numeric features are standardized using only the LAT-training portion of each outer fold, and Ridge regularization is fixed at $\alpha=1.0$. The penalty is fixed at this value for all reported models and is not tuned; we do not report a sweep over alternative penalties. The manifest and row-count inventory are part of the planned public reproducibility release.

\paragraph{Range sensitivity.} Ridge predictions are not bounded to the rating scale. We check the consequences on the predicted-quality-plus-delay model ($r=.475\pm.026$): clipping the predictions to the scale gives $r=.482\pm.022$, and a separately fitted bounded model gives $r=.463\pm.038$. The ordering of the systems in Section~5.2 is unaffected.

\paragraph{Delay-representation checks.} The alternative delay representations in Section~4.1 are evaluated with the same leave-one-speech-group-out protocol and the same fixed $\ell_2$ penalty as the delay-only model, using the two-evaluator mean LAT target. They are delay-only comparisons and do not involve the quality model, so they do not require the cross-fitted two-stage pipeline. We report them as a descriptive check on the delay basis rather than as part of the primary evaluation.

\paragraph{Range and calibration checks.} Because the target is bounded to $[.5,3.0]$ but Ridge is unbounded, we check the predicted-quality-plus-delay model for each seed. As sensitivity analyses we post-hoc clip the held-out prediction, $\mathrm{clip}(\hat y,.5,3.0)$, and fit a bounded model $\hat y=.5+2.5\,\sigma(X\beta)$ separately within each fold using the same fixed $\ell_2$ penalty; neither uses outer-test labels. The clipped and bounded Pearson correlations appear in Section~5.2. The out-of-range prediction counts, MSE, MAE, descriptive calibration from $y=a+b\hat y$, and a constant fold-training-mean reference are part of the reproducibility release; these checks assess range and calibration robustness rather than select a new primary model.

\paragraph{Held-out interpreter setup.} Every segment from the outer interpreter is excluded from quality training, development, checkpoint selection, and LAT-model training. Within the remaining interpreters, four inner speech-group partitions generate OOF predicted LQ/EXP for LAT-model training; a final interpreter-excluded quality model predicts the held-out interpreter. We compare delay-only, structural features with delay, predicted quality with delay, and their combination, with no interpreter identity or global-mean feature. Source speeches may still appear through other interpreters, so this protocol measures a pooled association after excluding an interpreter, not joint transfer to both held-out interpreters and held-out source speeches, and not uniform generalization to every new interpreter.

\paragraph{Bootstrap and permutation-test details.} Automatic point estimates are means over three fixed training seeds; their reported $\pm$ values are sample SDs across those seeds, not sampling confidence intervals. For source-speech-group-held-out results, we use 10,000 cluster-bootstrap draws: we resample the 16 source-speech groups with replacement, and each selected group contributes all of its segments. The identical group draw is applied to paired systems. Two conventions are used: Section~5.2 model and paired-difference intervals (e.g.\ $\Delta r=.040$) average per-seed metrics within each draw, matching the main-table convention; the full-minus-structural increment ($\Delta r=.030$) and the permutation test average seed predictions before the statistic. Pearson is nonlinear, so these differ.

For the primary construct comparison (predicted quality+delay versus timing-only), we enumerate all $2^{16}=65{,}536$ whole-group label swaps, preserve unequal observed group sizes in corpus correlation, use a two-sided alternative, and apply the plus-one correction
\begin{equation}
p =
\frac{
1 + \#\left\{ |T^{*}| \ge |T| \right\}
}{
1 + 65{,}536
}.
\end{equation}
This is the sole confirmatory test. Seed-wise swaps and held-out-interpreter results are descriptive. We report prediction SDs to detect collapse; the human-label model is a non-deployable rubric reference. Associations are predictive, not causal evidence about ratings.

\section*{Appendix C: Additional Analyses}

\paragraph{Rater agreement.} On all 622 jointly rated segments, R05 assigned scores
$\{0,1,2,3\}$ with counts $(19,72,186,345)$ and R06 with counts
$(0,33,187,402)$. The rows of Table~\ref{tab:rater-confusion} are R05 scores
and the columns are R06 scores. Exact agreement is 51.4\%; within-one-point
agreement is 91.5\%; Pearson and Spearman correlations are .259 and .261;
ICC$(2,1)$ and quadratic-weighted kappa are .236 and .236.

\begin{table}[h]
\centering
\small
\caption{LAT-score confusion matrix: R05 rows, R06 columns.}
\label{tab:rater-confusion}
\begin{tabular}{c|rrrr}
\toprule
R05 $\backslash$ R06 & 0 & 1 & 2 & 3 \\
\midrule
0 & 0 & 4 & 8 & 7 \\
1 & 0 & 3 & 41 & 28 \\
2 & 0 & 16 & 60 & 110 \\
3 & 0 & 10 & 78 & 257 \\
\bottomrule
\end{tabular}
\end{table}

\paragraph{Evaluator-specific results.} Table~\ref{tab:rater-human} uses human LQ and EXP ratings to examine how each evaluator relates quality to LAT. These results are used to study the rating behavior, not as deployable automatic systems.

R05 shows a much stronger within-evaluator relation between quality and LAT than R06. This difference supports the separate evaluator analysis in Section~5.4.

\begin{table}[h]
\centering
\scriptsize
\setlength{\tabcolsep}{2.5pt}
\caption{Evaluator-specific results using human LQ/EXP ratings. Models are fitted within the same outer speech-group folds; intervals are speech-group clustered 95\% CIs.}
\label{tab:rater-human}
\begin{tabular}{llccc}
\toprule
Target & System & Pearson $r$ & Spearman $r_S$ & MSE \\
\midrule
R05 & Delay & .106 [.014,.220] & .098 & .656 \\
R05 & Same-rater LQ+EXP+delay & .909 [.855,.944] & .836 & .113 \\
R05 & Structure+delay & .613 [.457,.691] & .546 & .406 \\
R05 & Full human reference & .911 [.859,.944] & .840 & .111 \\
R06 & Delay & .400 [.293,.488] & .403 & .293 \\
R06 & Same-rater LQ+EXP+delay & .552 [.496,.607] & .547 & .242 \\
R06 & Structure+delay & .442 [.373,.498] & .479 & .281 \\
R06 & Full human reference & .545 [.481,.610] & .552 & .245 \\
\bottomrule
\end{tabular}
\setlength{\tabcolsep}{6pt}
\end{table}

\paragraph{Residualized quality features.} For each training fold, we fit a model that predicts automatic LQ and EXP from the structural features. We then subtract these predictions to obtain quality residuals for both training and test data. Structural features explain $R^2=.200/.199$, $.246/.114$, and $.215/.144$ of LQ/EXP across the three seeds. Replacing the original quality scores with these residuals gives almost identical Pearson correlations to the full model, agreeing to three decimal places: $.665$, $.663$, and $.646$. The corresponding confidence intervals for the improvement over structure plus delay are [--.006,.073], [.004,.061], and [--.016,.046].

\paragraph{Results by language direction.} Table~\ref{tab:rater-direction}
reports the two most informative human-label systems by direction. The
English-to-Chinese condition contains only 117 segments, so these intervals are
descriptive rather than a direction-level model comparison.

\begin{table}[h]
\centering
\scriptsize
\setlength{\tabcolsep}{2.5pt}
\caption{Results by language direction using human LQ/EXP ratings. Brackets are speech-group clustered 95\% CIs.}
\label{tab:rater-direction}
\begin{tabular}{llrrr}
\toprule
Target & System & Direction ($n$) & Pearson $r$ & MSE \\
\midrule
R05 & LQ+EXP+delay & Zh$\rightarrow$En (505) & .895 [.814,.948] & .103 \\
R05 & LQ+EXP+delay & En$\rightarrow$Zh (117) & .929 [.860,.981] & .157 \\
R05 & Structure+delay & Zh$\rightarrow$En (505) & .508 [.262,.585] & .384 \\
R05 & Structure+delay & En$\rightarrow$Zh (117) & .740 [.663,.775] & .503 \\
R06 & LQ+EXP+delay & Zh$\rightarrow$En (505) & .537 [.466,.601] & .258 \\
R06 & LQ+EXP+delay & En$\rightarrow$Zh (117) & .641 [.464,.732] & .173 \\
R06 & Structure+delay & Zh$\rightarrow$En (505) & .428 [.342,.486] & .295 \\
R06 & Structure+delay & En$\rightarrow$Zh (117) & .466 [.372,.548] & .221 \\
\bottomrule
\end{tabular}
\setlength{\tabcolsep}{6pt}
\end{table}

\paragraph{Rating procedure details.} Evaluators listened to synchronized audio, viewed transcripts, could replay segments, and did not see the numerical delay values. The resulting judgment is therefore a deliberative one made with audio and transcripts, not purely auditory real-time perception. The archive does not preserve headphone-use records, rater-training or calibration-session records, presentation-order randomization, rater-independence confirmation, or score/comment-order records. We therefore cannot rule out session-order or environment-related effects beyond the halo-effect concern already discussed in the main text.

\paragraph{Reproducibility details.} The public reproducibility package will include the
corrected-data release tag and correction log, a table-to-file provenance map,
the exact outer and inner fold manifests, train/development/test counts,
environment versions, model-card configuration, range and calibration checks,
direction and per-interpreter tables, and the qualitative case-selection rule.

\section*{Appendix D: Illustrative Examples}

Table~\ref{tab:case-pairs} shows one case with a high LAT rating and one with a low rating. The examples were selected after the quantitative analysis to illustrate why onset delay alone does not always match professional judgment. The comments are shortened and translated where necessary and are never used as model inputs.

\begin{table*}[t]
\centering
\small
\renewcommand{\arraystretch}{1.12}
\setlength{\tabcolsep}{4pt}
\caption{Observed professional-comment examples. These are illustrations, not additional rubric anchors or quantitative evidence.}
\label{tab:case-pairs}
\begin{tabular}{p{.08\textwidth}p{.52\textwidth}p{.32\textwidth}}
\toprule
Case & Source and interpreted output & Scores and professional comment \\
\midrule
Positive & \textbf{Source:} An English conditional explains that many children in orphanages are not orphans, so ``orphanage'' is a euphemistic label. \newline \textbf{Interpreted output:} The Chinese output retains both the condition and conclusion. & LAT $3.0$; predicted $3.05$; delay 2.00~s. \newline ``Fast, steady, complete, and idiomatic.'' \\
Negative & \textbf{Source:} An English segment states that a Mars trip requires an Earth--Mars alignment launch window. \newline \textbf{Interpreted output:} The Chinese output gives an incorrect distance and omits the launch-window condition. & LAT $.5$; predicted $2.35$; delay 2.44~s. \newline ``Limited information capture; unable to form complete sentences.'' \\
\bottomrule
\end{tabular}
\renewcommand{\arraystretch}{1}
\setlength{\tabcolsep}{6pt}
\end{table*}

\section*{Appendix E: Structural Feature-Family Ablation}
\label{app:ablation}
This appendix gives the detailed ablation summarized in Section~5.3. All configurations retain the delay basis $\phi(d_s)$, and the cohort, folds, standardization, and Ridge penalty are unchanged from Section~4.4. Feature families are length (source and output length), ratio (output-to-source length ratio), punctuation (punctuation and sentence-ending counts), diversity (lexical diversity), very-short (the very-short-output indicator), and direction (the language-direction indicator).

Point estimates are observed differences on the retained cohort. Intervals are 95\% percentile intervals of the same difference over 10,000 bootstrap resamples of the 16 source-speech groups, drawn with replacement, keeping all segments of a sampled group together. These intervals are exploratory, not the confirmatory permutation test of Section~5.2.

\begin{table}[t]
\centering
\footnotesize
\setlength{\tabcolsep}{3pt}
\caption{Structural feature-family ablation. Panel A adds one family to onset delay; Panel B removes one family from the full structural set. In Panel A, $\Delta r = r - r(\text{delay only})$, so a positive value is the gain from adding the family. In Panel B, $\Delta r = r(\text{full}) - r$, so a positive value is the loss from removing it. Starred differences have bootstrap intervals excluding zero.}
\label{tab:family-ablation}
\begin{tabular}{lrrrrr}
\toprule
Feature set & $k$ & $r$ & $r_S$ & MSE & $\Delta r$ \\
\midrule
\multicolumn{6}{l}{\emph{Panel A: added to onset delay}} \\
Onset delay only & 5 & .318 & .347 & .281 & --- \\
\quad $+$ length & 7 & .486 & .517 & .241 & $+.168$* \\
\quad $+$ length ratio & 6 & .342 & .375 & .277 & $+.024$ \\
\quad $+$ punctuation & 7 & .310 & .352 & .284 & $-.008$ \\
\quad $+$ lexical diversity & 6 & .347 & .379 & .275 & $+.029$ \\
\quad $+$ very-short & 6 & .318 & .347 & .281 & $.000$ \\
\quad $+$ direction & 6 & .304 & .328 & .285 & $-.014$ \\
\midrule
\multicolumn{6}{l}{\emph{Panel B: removed from the full set}} \\
Structural $+$ delay (full) & 13 & .632 & .618 & .187 & --- \\
\quad $-$ length & 11 & .584 & .603 & .205 & $+.048$* \\
\quad $-$ length ratio & 12 & .634 & .620 & .186 & $-.002$ \\
\quad $-$ punctuation & 11 & .639 & .624 & .184 & $-.007$* \\
\quad $-$ lexical diversity & 12 & .631 & .615 & .188 & $+.002$ \\
\quad $-$ very-short & 12 & .632 & .618 & .187 & $.000$ \\
\quad $-$ direction & 12 & .491 & .512 & .240 & $+.141$* \\
\quad $-$ length blk., ratio kept & 9 & .559 & .578 & .214 & $+.074$* \\
\quad $-$ length blk. and ratio & 8 & .290 & .329 & .290 & $+.342$* \\
\bottomrule
\end{tabular}

\vspace{2pt}
{\footnotesize Starred intervals, in row order: $+$length $[.052,.268]$; $-$length $[.009,.078]$; $-$punct.\ $[-.012,-.003]$; $-$direction $[.029,.247]$; $-$length blk.\ $[.009,.128]$; $-$length blk.\ and ratio $[.149,.474]$. Full against delay only: $+.314$ $[.134,.426]$. Unstarred intervals include zero: $+$ratio $[-.026,.094]$, $+$punct.\ $[-.065,.029]$, $+$div.\ $[-.030,.066]$, $+$direction $[-.095,.041]$, $-$ratio $[-.007,.003]$, $-$div.\ $[-.007,.009]$.\par}
\setlength{\tabcolsep}{6pt}
\end{table}

\paragraph{Interpretation limits.} The correlations behind the feature-group reading of Section~5.3 are as follows. Output length correlates with the length ratio at $.85$, with direction at $.76$, with lexical diversity at $-.73$, and with punctuation at $.68$; the ratio correlates with direction at $.78$, and the two punctuation features at $.72$. The delay basis is collinear by construction ($.99$ for delay and its first hinge). Two consequences follow: the row removing the length block and ratio is not a length-free condition, since punctuation counts still covary with output length; and a family whose removal leaves performance unchanged may still be informative, since the remaining features may carry the same information.

\paragraph{Very-short-output indicator.} The shortest retained output has exactly five lexical units, so the indicator is constant at zero with a zero Ridge coefficient and leaves every metric in Table~\ref{tab:family-ablation} unchanged to three decimals. The cohort contains no very short outputs, so their relevance to promptness is untested here rather than disconfirmed.

\paragraph{Parsimony diagnostic.} A reduced set of source length, output length, and the direction indicator, with delay, reaches $r=.640$ against $.632$ for the full structural set ($\Delta r=+.007$, $[.002,.017]$). This diagnoses where the signal concentrates and is not a proposed model: the difference is small, the interval exploratory, and the reduced set is not consistently ahead across groups. All main structural and full-model results use the full structural feature set.

\begin{table}[t]
\centering
\footnotesize
\caption{Evaluator-specific LAT prediction with automatic features, as summarized in Section~5.4. The cohort, folds, features, and regularization are identical across rows; only the target changes. $\Delta r$ is the structural gain over delay only, with an exploratory 95\% speech-group bootstrap interval. MSE is not comparable across targets because the target standard deviations differ.}
\label{tab:rater-auto}
\begin{tabular}{lccccc}
\toprule
 & \multicolumn{2}{c}{Delay only} & \multicolumn{2}{c}{Struct.\ $+$ delay} & \\
\cmidrule(lr){2-3}\cmidrule(lr){4-5}
Target & $r$ & $r_S$ & $r$ & $r_S$ & $\Delta r$ \\
\midrule
Mean & .318 & .347 & .632 & .618 & $+.314$ \\
R05 & .106 & .098 & .613 & .546 & $+.507$ \\
R06 & .400 & .403 & .442 & .479 & $+.042$ \\
\bottomrule
\end{tabular}

\vspace{2pt}
{\footnotesize Target SDs: mean $.558$, R05 $.806$, R06 $.589$. MSE, delay only: $.281$, $.656$, $.293$; structural $+$ delay: $.187$, $.406$, $.281$. $\Delta r$ intervals: mean $[.134,.426]$, R05 $[.329,.618]$, R06 $[-.034,.123]$. The difference between the R05 and R06 improvements was not tested directly, so no statistically significant between-evaluator difference is established.\par}
\end{table}

\paragraph{Language direction.} Every speech group is single-direction (ten Chinese-to-English, 505 segments; six English-to-Chinese, 117), so direction is constant within each group and its contribution cannot be cleanly separated from speech-level differences in topic, speaker, or interpreter. The directions also differ in output-length scale ($113.0$ output units and ratio $1.29$ for English-to-Chinese against $34.4$ and $0.50$), so the additive indicator plausibly helps adjust for this systematic difference between directions; we treat this as a conjecture, not an established mechanism.

\end{document}